\documentclass[12pt]{article}
\usepackage[utf8]{inputenc}
\usepackage{amsmath}
\usepackage{amssymb}
\usepackage{bbold}
\usepackage[title]{appendix}
\usepackage{comment}
\usepackage{float}
\usepackage{array}
\usepackage{graphicx}
\usepackage{hyperref}
\hypersetup{
    colorlinks=true,
    linkcolor=blue,
    filecolor=magenta,
    urlcolor=cyan,
}
\usepackage[natbibapa]{apacite}

\title{AI and Generative AI for Research Discovery and Summarization}
\author{Mark E. Glickman \\
Department of Statistics\\
Harvard Univeristy\\
glickman@fas.harvard.edu \thanks{
Department of Statistics, Harvard University, Cambridge, MA, USA, 02138} 
\and Yi Zhang \\
Department of Statistics\\
Harvard University \\
yi$\_$zhang@fas.harvard.edu
}
\date{}

\begin{document}

\maketitle

\begin{abstract}
AI and generative AI tools, including chatbots like ChatGPT that rely on large language models (LLMs), have burst onto the scene this year, creating incredible opportunities to increase work productivity and improve our lives.  Statisticians and data scientists have begun experiencing the benefits from the availability of these tools in numerous ways, such as the generation of programming code from text prompts to analyze data or fit statistical models.  One area that these tools can make a substantial impact is in research discovery and summarization.  Standalone tools and plugins to chatbots are being developed that allow researchers to more quickly find relevant literature than pre-2023 search tools.  Furthermore, generative AI tools have improved to the point where they can summarize and extract the key points from research articles in succinct language.  Finally, chatbots based on highly parameterized LLMs can be used to simulate abductive reasoning, which provides researchers the ability to make connections among related technical topics, which can also be used for research discovery.  We review the developments in AI and generative AI for research discovery and summarization, and propose directions where these types of tools are likely to head in the future that may be of interest to statistician and data scientists.
\end{abstract}

\noindent {\bf Keywords}: Abductive reasoning,
Hallucination,
Literature discovery,
Manuscript abstraction,
Research discovery

\addtolength{\baselineskip}{12pt}

\newpage
\section{Introduction}\label{sec:intro}

In the past couple of years, the introduction of artificial intelligence (AI) 
tools into mainstream use has made an enormous impact on the way people work, learn,
and improve their lives.
This is particularly true given the widespread availability of generative AI
chatbots, such as Open AI's ChatGPT Plus and Microsoft's Copilot, both of which use
the GPT-4 \citep{openai2023gpt4} large language model (LLM), 
and Google Bard which relies on the Gemini Pro LLM.
For people who work in quantitative fields, the benefits are remarkable.
ChatGPT Plus can generate and run python code natively
when prompted to carry out a coding task.
This includes the ability to fit statistical models and produce visualizations.
It can also connect with other computing engines, such as Wolfram's Alpha, to 
perform computational tasks and analyses.
These chatbots can also be used as impressive learning tools and informal teachers,
with the ability to summarize unfamiliar topics in a much more
focused way than typical web searches.
LLMs are particularly well-suited to language translation (e.g., the 
Google Translate app), so that quantitative academic researchers who do not 
have English as a first language
now have an assistive tool to help translate their non-English writing into English. 
The scope of AI technology is advancing so quickly, it is difficult to keep track of the areas of application and of the tools becoming available.  As an example, ChatGPT Plus
established 3rd-party plugins in March 2023, with a total of 11 plugins \citep{whatplugin}, and has since grown to 1037 plugins as of mid-December 2023 \citep{scriptbyai}.

One of the most important ways for practicing statisticians and data scientists to remain viable in their work is being able to learn and apply unfamiliar methods.  Doing so requires the ability to perform literature searches, become familiar with quantitative subfields, and learn about methodological advances.  
Until relatively recently, statistical researchers have relied mostly on digital libraries such as JSTOR (\url{www.jstor.org}), which has for decades archived influential journals in statistics, including Journal of the American Statistical Association (JASA), Journal of the Royal Statistical Society - Series B (JRSS-B), Biometrika, the Annals of Statistics, and many more. However, with options for publication that go way beyond the JSTOR collection, other avenues have opened up. Many statistical researchers have relied on Google Scholar searches (\href{https://scholar.google.com}{scholar.google.com}) to find relevant articles based on provided keywords.  Based on key word input, Google Scholar ranks candidate documents using criteria such as the relevance of the work, where it was published, and the degree to which the work has been cited by other documents \citep{beel2009google}.  One of the key limitations of Google Scholar, however, is the lack of a LLM front end to interpret nuanced inputs.  

\begin{figure}[h!]
    \centering
    \includegraphics[width=0.9\textwidth]{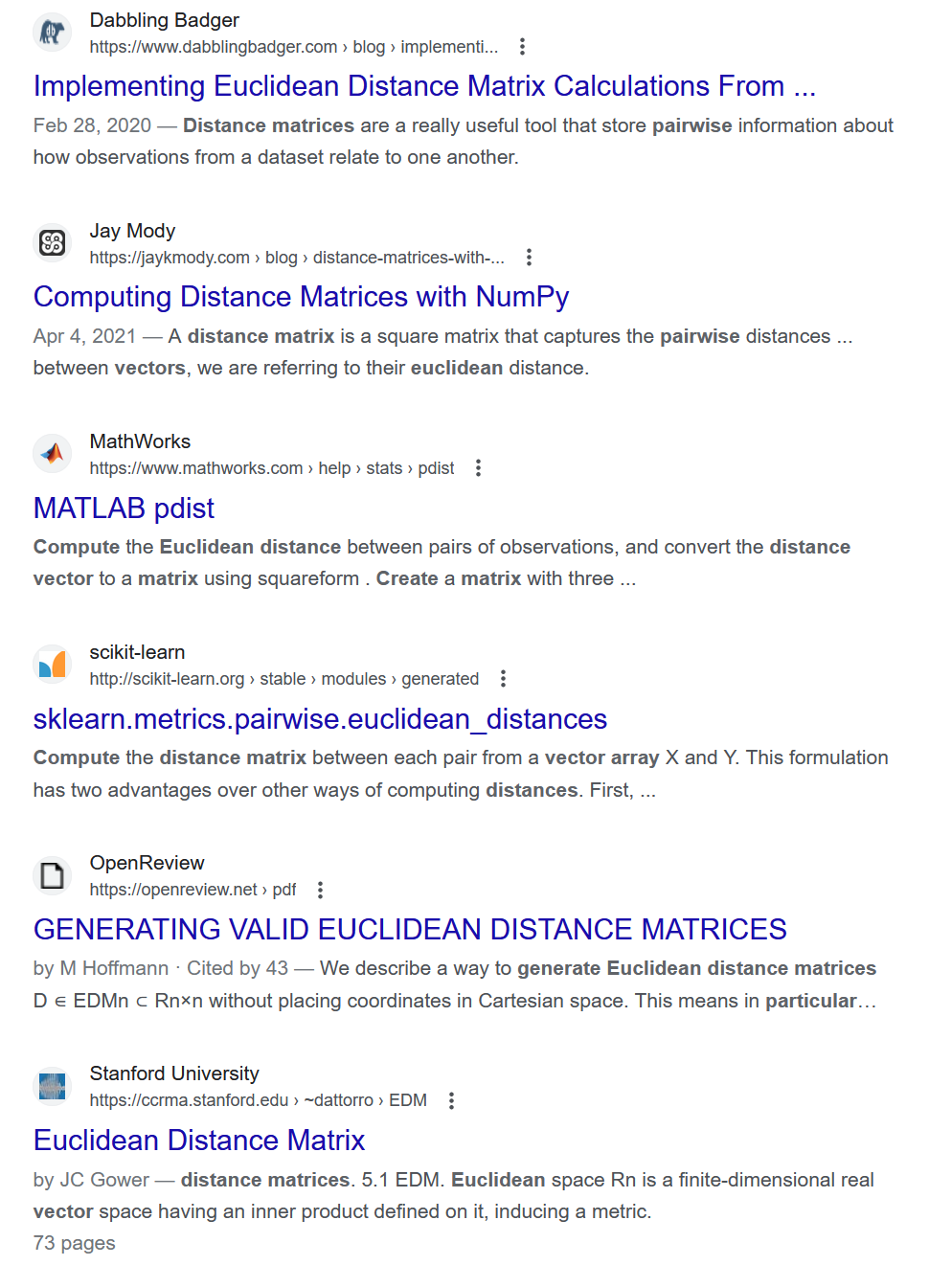}
    \caption{Google search results based on the keywords ``generating
    Euclidean vectors from a pairwise distance matrix.''}
    \label{fig:google-search}
\end{figure}

The following example, which we return to throughout this manuscript, demonstrates a limitation of non-LLM search tools.  Consider the problem in which a pairwise distance or dissimilarity matrix of $n$  objects has been computed, and it is of interest to generate $n$ vectors each of specified dimension $d$ (that is, $d$-dimensional embeddings) whose pairwise Euclidean distances correspond, at least approximately, to the initial pairwise distance matrix.  Such a procedure is known as classical multidimensional scaling \citep{torgerson1952multidimensional, gower1966some}, and is more commonly known among psychometricians than statisticians and data scientists.  
Performing a Google search using the keywords “generating Euclidean vectors from a pairwise distance matrix” returns the search results in Figure~\ref{fig:google-search}.

In every search result item on the first page (and several subsequent pages), Google returned links to methods for computing pairwise distance matrices based on Euclidean vectors, which is the opposite of what was requested.  Similarly, a Bing web search resulted in similar sets of results on computing distance matrices from vectors, though Bing did provide a link for a StackExchange question-and-answer mentioning the creation of vectors from a distance matrix, but without providing the correct answer \citep{converting2013}. 

The difficulty with this particular type of search query is that the parser may not be able to distinguish whether the query is about computing distance matrices or about determining vectors that produce given distance matrices.  Additionally, the search engine may not be able to distinguish whether web pages are addressing the question of interest or a question that involves a reconfiguration of the words in the prompt. 
Given that many more sites on the web are concerned with computing distance matrices from vectors, 
a search engine is more likely 
to produce results that address the more commonly queried topic.  
The role of AI in this type of search is to improve making sense of the question’s intent, and to have a more concrete “understanding” of website content.

This article reviews the current landscape of AI tools available to statistical and data science researchers to perform literature searches, to synthesize topics from disparate sources, and to improve the acquisition of knowledge for a data science audience.  
We describe some of the basic functions of existing tools, but our emphasis is on a high-level explanation of how AI tools can enhance work. 
We introduce in Section~\ref{sec:hallucinations} the problem that
chatbots sometimes provide fake responses
(``hallucinations'') to prompts, 
which can make some of their output unreliable.
Fortunately, recent improvements have been making chatbots less prone to hallucinating.
This is followed in Section~\ref{sec:abductive} by an examination of 
an important capability of
chatbots, specifically how they can be used to identify research methods simply 
by describing the workings of the methods.
This is a form of abductive reasoning, and LLM chatbots appear to perform 
(or rather, simulate) this type of reasoning well.
We then describe 
a sampling of tools that are popular for conducting informed literature discovery
(Section~\ref{sec:literature})
and manuscript summarization (Section~\ref{sec:summarizing}).
The tools that are available are constantly updating and improving, 
so it is only possible to provide a snapshot in time of their basic functioning.  
We conclude in Section~\ref{sec:discuss} with our hypotheses about how developments in AI tools for data science research will evolve, 
given the current state of technology.

\section{From web searches to chatbot queries: Hallucination issues}\label{sec:hallucinations}

Prior to the public release of various generative AI chatbots,
web searches have been the standard for information retrieval, offering results based on indexed web pages. However, tools like ChatGPT, powered by LLMs, represent a new paradigm. They provide conversational and context-aware responses, which is a departure from the list of links and summaries typical of standard web searches. Despite this innovative approach, ChatGPT and similar AI tools do face a unique challenge not commonly encountered in traditional web searches: the issue of ``hallucination.'' In the context of AI, hallucination typically refers to the generation of information that is either factually incorrect, irrelevant, or nonsensical, yet is often delivered with a high degree of confidence. This presents a unique challenge in ensuring the reliability and accuracy of the information provided by these AI systems.

As widely covered in news reports \citep[e.g.,][]{newyyorktimes1} and detailed in technical papers \citep[e.g.,][]{openai2023gpt4}, LLM-based chatbots, including ChatGPT, consistently encounter hallucinations. This issue goes beyond simple technical glitches, striking at the fundamental principles of how these models are trained and function. The comprehensive survey paper by \citet{ji2023survey} offers an in-depth examination of this phenomenon. It presents a detailed analysis of the various aspects and implications of hallucinations in language models, providing a comprehensive overview of how and why hallucinations occur in AI systems. 

A particularly notable case of hallucination in ChatGPT, which has been observed in multiple instances, is its tendency to fabricate references or cite non-existent sources. For example, in a study investigating the frequency of AI hallucinations in research proposals generated by ChatGPT, it was discovered that out of 178 references cited, 69 did not have a Digital Object Identifier (DOI), and 28 of these were found to be completely non-existent \citep{alkaissi2023artificial}. Similarly, another study evaluating the quality of the answers and the references provided by ChatGPT for medical questions 
found that, out of the 59 references included in the primary analysis, 41 (69\%) were fabricated. Moreover, among the remaining 18 valid references, several had 
other types of issues: 3 contained minor citation inaccuracies, and 5 had major citation errors \citep{gravel2023learning}.

\begin{figure}[h!]
    \centering
    \includegraphics[width=1.0\textwidth]{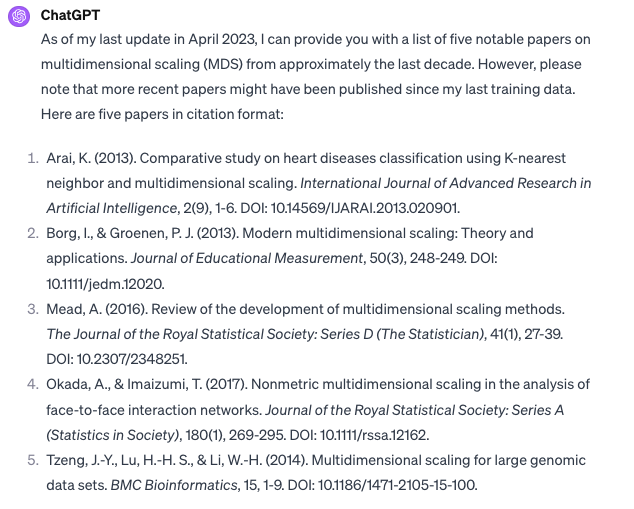}
    \caption{Response of ChatGPT to the prompt ``provide citations for five papers on multidimensional  scaling from the last ten years.''}
    \label{fig:hallucination}
\end{figure}

Such hallucinations are not the exception, but rather the norm. We requested ChatGPT to provide five recent papers on ``multidimensional scaling'' from the past decade. The search results are presented in Figure~\ref{fig:hallucination}. While these results initially seem credible, a closer look found that two of the five papers (citations 1 and 4) listed actually do not exist. Moreover, among the three valid references, each contains citation inaccuracies, including nonexistent DOIs, incorrect publication years, and in one case (citation 2), an incorrect journal name.

This pattern of hallucinations in ChatGPT's output significantly undermines the reliability of AI-generated information, especially when used as a search tool for academic content where precision and factual accuracy are not merely preferred, but essential. Consequently, while LLMs like ChatGPT can augment traditional search methods by providing quick, comprehensive, and sometimes insightful summaries, their use must be complemented with an appreciation of their limitations. 

Current efforts are focusing on developing more sophisticated training methods, such as using verified and fact-checked datasets, and implementing advanced algorithms capable of cross-referencing and validating information against trusted sources.
Innovations in this area include tools that enable LLMs to autonomously verify information by interfacing with trusted databases (e.g., databases of peer-reviewed articles), which significantly reduce the incidence of hallucinations by ensuring the accuracy of the generated content.
To this end, chatbots like ChatGPT Plus 
have incorporated plugins that specifically search for scholarly content and grab the actual publication details rather than let the LLM hallucinate document details on its own.  Moreover, there is a growing recognition of the importance of human oversight in the use of LLMs. Integrating expert review 
can provide an additional layer of verification. Researchers and users are best served by approaching the LLM output with a critical eye, verifying against primary sources and employing a combination of methods for thorough research.

\section{Method identification: Abductive reasoning} \label{sec:abductive}

A research-related task that does not ask a chatbot
to provide valid references is in the area of method identification
and topic discovery.
A researcher may have an idea for a novel approach to a problem, and wants 
to know whether such an approach already exists and has been developed 
and studied. 
Alternatively, the researcher may want to learn 
about related methods to address a problem.
Another situation is where a researcher may suspect that an algorithm, statistical procedure, or computational method may already exist, but is unaware of the conventional name of the procedure.
Addressing these tasks are instances of abductive reasoning.

Abductive reasoning is the
process of inferring the most believable explanation based on a given set of observations or statements \citep{peirce1935collected, peirce1973lectures}.  Typically, potential explanations may not be spelled out a priori, so unlike statistical inferential settings in which a set of potential explanations is usually listed, some notion of creativity is typically associated with the abductive process.  A classic example of abductive reasoning is when a patient mentions to their medical provider various symptoms from which they are suffering.  The provider then needs to (abductively) reason the likeliest explanation of the symptoms before proposing treatment.
In the above example with the researcher wanting to ascertain
whether an analytic approach already exists and has a conventional name,
the details of the algorithm (the ``observations'') are provided to the chatbot, 
and then the conventional description that most accurately describes the 
process (the ``explanation'') is identified by the chatbot in its simulation
of abductive reasoning.

These abductive research discovery tasks are now being successfully carried out using LLMs.  It is a remarkable feat that LLMs can mimic various aspects of logical reasoning, despite not being trained to do so.  Evidence from recent work \citep{wei2022emergent, wei2022chain, suzgun2022challenging} suggests that the more highly parameterized the LLM, the greater the likelihood for the LLM to exhibit reasoning ability in different forms.  Examination of the ability of LLMs to reason successfully have produced mixed results \citep{qiao2022reasoning, guo2023how, huang2022towards, davis2023mathematics}, though many of the documented conclusions may already be out of date.  \citet{bang2023multitask} assessed ChatGPT's abductive reasoning ability based on a challenge dataset \citep{bhagavatula2019abductive} that tests abductive natural language inference.  The authors noted that ChatGPT achieved 86.7\% accuracy, and that ChatGPT performed better at deductive and abductive reasoning than inductive reasoning.  \citet{pareschi2023abductive} has shown that advanced LLMs like GPT-4 show promise in abductive reasoning in complex situations.

We have found that chatbots such as ChatGPT Plus can be effective abductive reasoners to discover existing methods based on descriptions.  As a simple example, suppose we are interested in learning about methods that can be used to compare the distribution of a binary variable across different groups, as an analog to ANOVA for quantitative outcomes.  Using the prompt ``What methods exist for testing whether the distribution of a binary variable across different groups is the same, like how ANOVA is often used for quantitative outcomes?'', ChatGPT responds with a summary of common methods that include brief descriptions of chi-square tests, Fisher's Exact test, logistic regression, the Mantel-Haenszel test, and a few others that are also relevant.  For a researcher who may not have been familiar with these standard approaches, ChatGPT's response provides an answer that can be used as a starting point for follow-up.

The earlier example on generating embeddings given a pairwise distance matrix is even more compelling, especially given the failure of commonly used search engines to direct the user to appropriate websites.  When prompted with, ``How do I generate Euclidean vectors from a given pairwise distance matrix?  Explain the basic approach,'' ChatGPT responds with ``Generating Euclidean vectors from a given pairwise distance matrix is a task that involves multidimensional scaling (MDS). The goal is to find a set of points in a Euclidean space such that the distances between points in this space are as close as possible to the distances given in the pairwise distance matrix.'' It then provides the correct algorithm to construct the embeddings.

\section{Literature discovery} \label{sec:literature}

Literature discovery and comparative analysis of research literature often come up as some of the most time-consuming areas when conducting research. As we discuss below, generative AI has been making a big impact in these areas. 

\subsection{Standalone web-based literature search tools}\label{subsec:standalone}

A literature search engine with a similar purpose to Google Scholar and  which
serves as the back-end for several AI-enhanced tools is Semantic Scholar (\url{www.semanticscholar.org}).  This free literature search tool is popular with developers given its reliable API. The Semantic Scholar engine, which focuses on scientific literature searches, has the same limitations on input as Google Scholar, but relies more heavily on citation networks and collaborative filtering to identify relevant documents.  It can also fine-tune literature searches by allowing the user to specify which documents in its initial search were relevant and which were not, allowing better results after iterating the search.  While not strictly an AI tool on its own, Semantic Scholar, along with other recent literature search tools we describe below, has begun enhancing its utility by adding generative AI features.  For example, a search on Semantic Scholar accompanies many suggested documents with a brief, typically one-sentence, summary of the contents of the work.  These distillations are created using GPT-3 \citep{brown2020LanguageMA} style parsing of an article's abstract, body, conclusion and title \citep{cachola2020tldr}.  At the time of writing, Semantic Scholar has a database of over 200 million papers over which searches can be performed.

Over the past two years, many standalone web-based literature search
tools have entered the market.
The main distinction between these tools and ones like Google Scholar and Semantic Scholar is the use of LLMs to interpret the research prompt provided by the user.  This feature may be used to improve the relevance of search queries, or provide downstream output based on the form of the prompt.  One of the most popular tools is Consensus (\url{https://consensus.app}).  Early in 2023, the engine had a fairly restricted database of articles that focused on only six areas of scientific inquiry, but more recently Consensus has expanded to the Semantic Scholar database.  
When a user makes a query,
Consensus searches titles and abstracts of all the documents in its database
for the non-stop words within the prompt.
If the user asks a question as their prompt, Consensus provides
an LLM-generated answer based on the results of the search.

Another popular standalone tool is Assistant by Scite (\url{https://scite.ai/assistant}).  Like Semantic Scholar, Scite's Assistant performs a key word search to generate candidate works relevant to the search query.  The main strength of Scite is its ability to sort the results by a more complex citation algorithm \citep{nicholson2021scite}.  Because Scite is focused on citation networks and algorithms, the search tool can provide forward citations of articles on the search list results as easily as it can construct traditional backward citations.  Much of the use for Scite is on finding articles that may support or contrast the search prompt.  This may be more appropriate to fields with domain-specific queries, but probably not so directly useful for statistical or data science research.

A standalone web-based literature search tool that we explore below in more detail is Elicit \citep{kung2023elicit}, which can be accessed at \url{https://elicit.com/}.  Elicit also uses Semantic Scholar as its database, and finds relevant documents that are semantically related to the research question.  The summaries from Elicit include one-sentence summaries of each suggested work (like those directly produced by Semantic Scholar), along with a GPT-3 created overall paragraph summary of the top papers in the list.  The user can interact with individual articles by clicking on them, revealing more information about the article.  A useful feature is the ability to view critiques of the work (determined through a semantic analysis) by forward-cited articles.  

\begin{figure}[h!]
    \centering
    \includegraphics[width=1.1\textwidth]{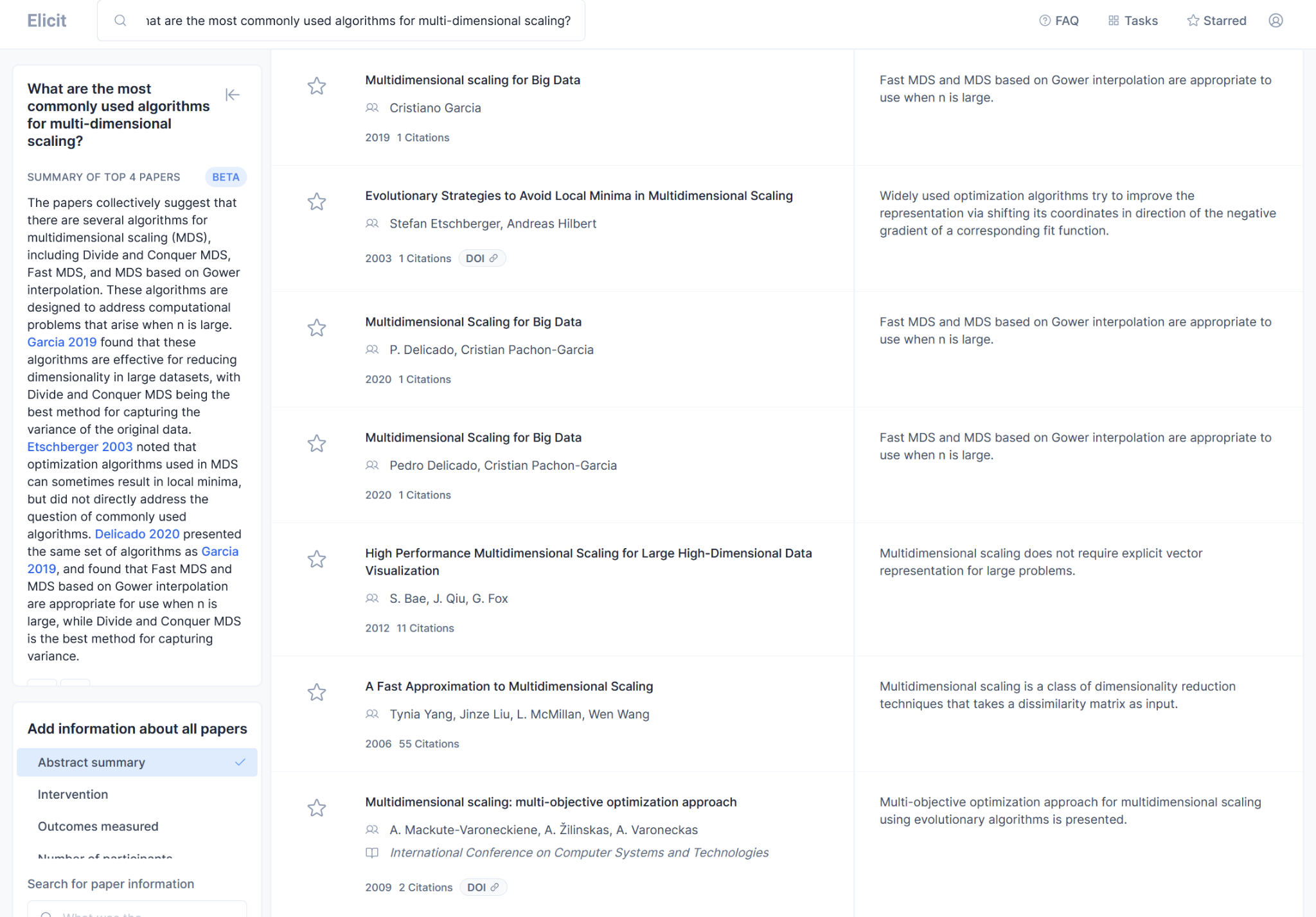}
    \caption{Results of query to Elicit.}
    \label{fig:elicit1}
\end{figure}

As an example of the use of Elicit, we asked it, ``What methods can be used to perform multidimensional scaling based on a pairwise distance matrix?'' The results were relevant, but most were older references.  We used a now-deprecated feature of Elicit, called ``brainstorming'', to generate the useful follow-up question ``What are the most commonly used algorithms for multidimensional scaling?''  This produced a more useful set of articles, the results of which can be seen in Figure~\ref{fig:elicit1}. 

\begin{figure}[h!]
    \centering
    \includegraphics[width=1.1\textwidth]{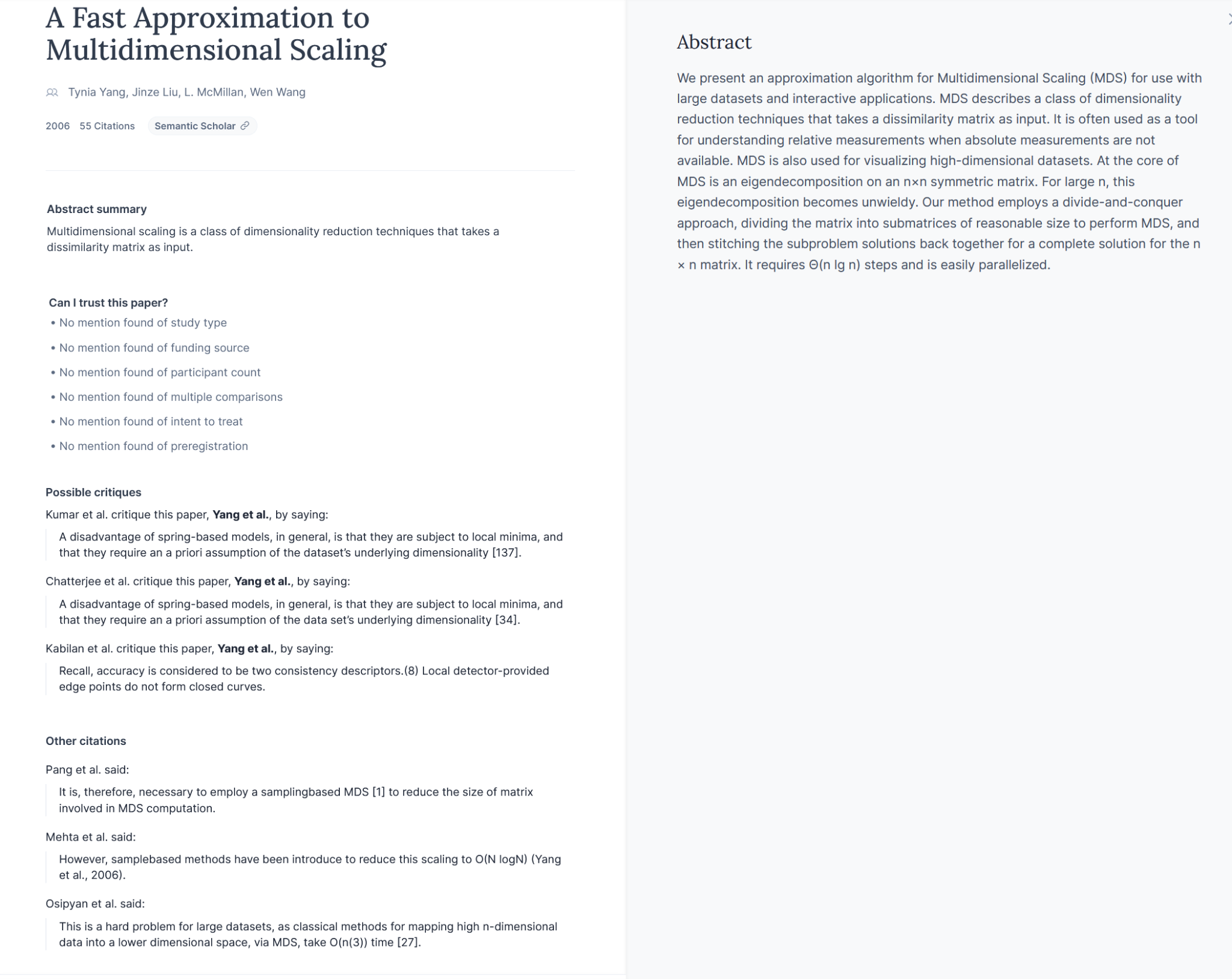}
    \caption{Information provided by Elicit on selected article.}
    \label{fig:elicit2}
\end{figure}

The article in the list by \citet{yang2006fast} seemed to be worthy of follow-up, and upon selecting that article, we obtained the summary seen in Figure~\ref{fig:elicit2}. 
Again, we note that in addition to the abstract summary, the possible critiques and other citations can help in evaluating the utility
of the referenced article.
These standalone tools have other helpful features, such as the ability to further explore the text content of suggested articles via natural language processing (NLP) analyses, export articles and meta-data, and to save and organize previous sessions.  It should be noted that these standalone web-based tools are moving in the direction of being fee-based, and the feature selections described here may be out of date as the development of these tools quickly evolve.

\subsection{Standalone web-based literature mapping tools}\label{subsec:mapping}

AI-powered engines like Litmaps (\href{https://www.litmaps.com/}{www.litmaps.com}) and ResearchRabbit (\href{https://www.researchrabbit.ai/}{www.researchrabbit.ai}) simplify the challenging task of identifying potential gaps in existing literature by visualizing the connections among different academic publications, helping to highlight overlooked or underexplored study areas. 
Litmaps is an innovative tool designed to streamline the process of literature review for researchers and academics. Utilizing citation searches tailored to databases, Litmaps reveals papers related to a specific topic and then visually maps out the relationships among these studies, providing a cohesive overview of the scientific landscape. This interactive tool generates literature maps that not only help researchers find relevant papers and understand the connections among them, but also facilitate the creation of a customized repository to aid in the literature review process.

\begin{figure}[h!]
    \centering
    \includegraphics[width=1.1\textwidth]{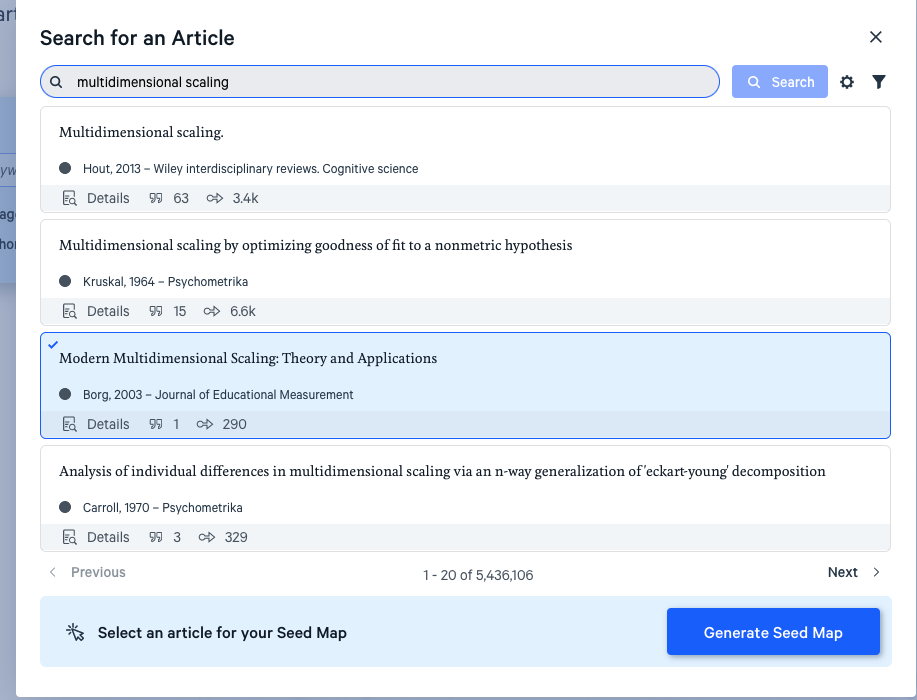}
    \caption{Results of search in Litmaps based on keywords.}
    \label{fig:litmaps1}
\end{figure}

\begin{figure}[h!]
    \centering
    \includegraphics[width=1.1\textwidth]{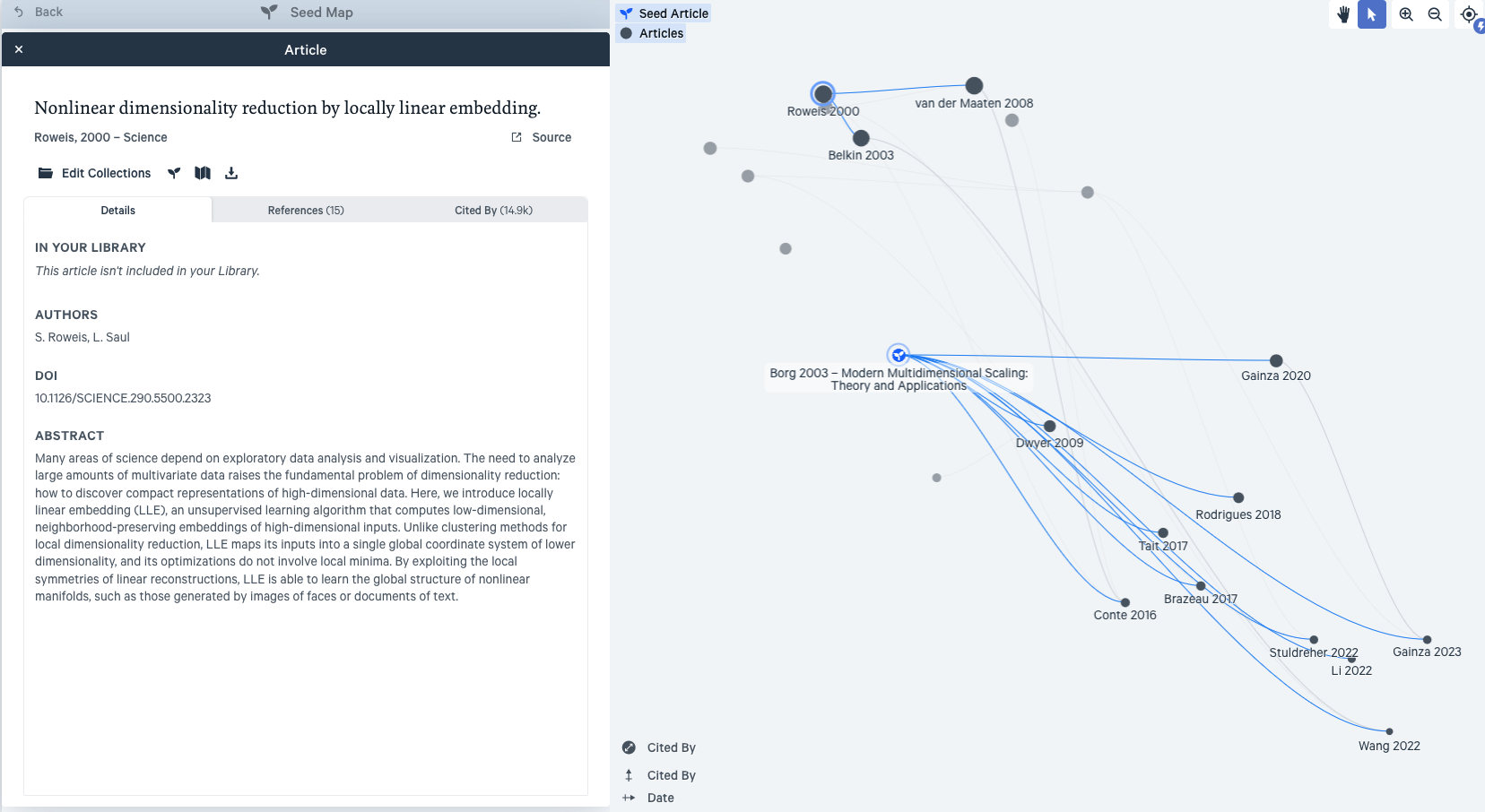}
    \caption{A display of the relevant literature on ``local linear embedding''
    based on the generated seed map.}
    \label{fig:litmaps2}
\end{figure}

As a simple application, we return to our previous example of generating embeddings from a distance matrix. By inputting the keywords ``multidimensional scaling'', Litmaps 
displays a list of relevant articles and papers. The user can then mark preferred articles and proceed to click on the ``Generate Seed Map'' button to initiate a literature review, as shown in Figure~\ref{fig:litmaps1}. 
To illustrate, we selected the published book \citet{borg2005modern} as a starting point, at which point Litmaps then generates a seed map, finding papers that either cite or are cited by this foundational piece. In the top left corner of the map, the paper on ``Local Linear Embedding'' \citep{roweis2000nonlinear} is displayed as seen in Figure~\ref{fig:litmaps2}, which has received numerous citations reflecting its impact in dimension reduction. 

\begin{figure}[h!]
    \centering
    \includegraphics[width=1.1\textwidth]{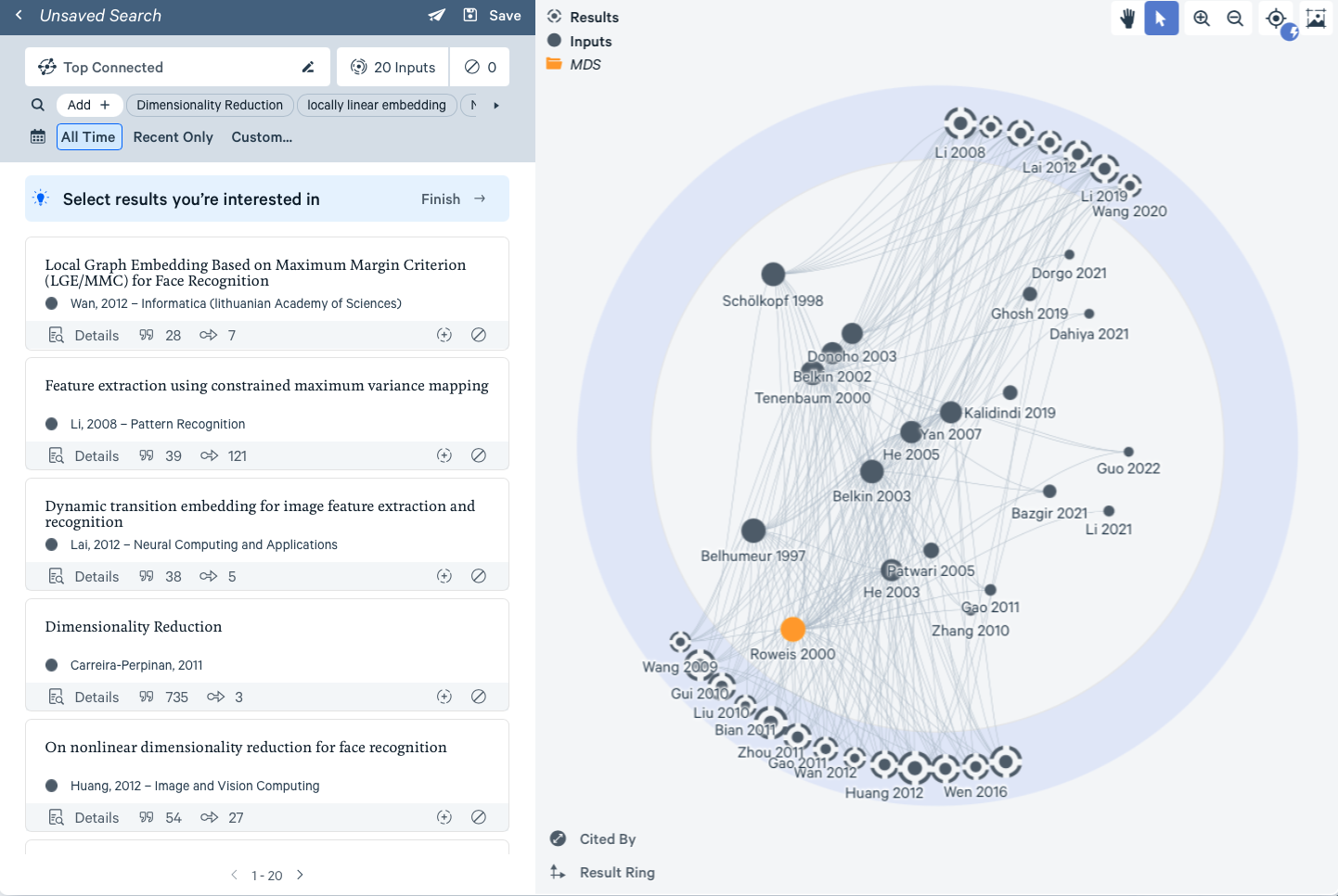}
    \caption{Results of the ``Discovery'' feature in Litmaps.}
    \label{fig:litmaps3}
\end{figure}

Building upon this newly selected article and its associated seed map, the “Discovery” feature in Litmaps now accommodates multiple seed inputs, piecing together an extensive discovery map that encompasses a wider array of related literature. This map is organized such that our initial selections form the core in the inner circle, while the outer circle highlights the most relevant additional works. The results are illustrated in Figure~\ref{fig:litmaps3}.

These visualization tools, serving as a kind of ``Spotify'' for academia, have significantly expanded the possibilities for literature discovery, especially in identifying papers that cross multiple topics or disciplines. They simplify the process of finding related papers and enhance the ability to efficiently access a broad spectrum of academic knowledge. Importantly, these tools can also uncover hidden, cross-disciplinary connections that might be missed when specialists concentrate exclusively on their own fields.

\subsection{Within-ChatGPT plugin tools/Custom GPTs}  \label{subsec:plugins}

In addition to standalone applications, ChatGPT plugins have extended the functionality of the base ChatGPT chatbot,
enabling it to perform specialized tasks related to research discovery and summarization.

One popular tool is the ScholarAI plugin (\href{https://scholarai.io/}{scholarai.io}), which is tailored to help researchers access a wide range of peer-reviewed articles, journals, and conference papers. It simplifies the process of finding relevant literature by allowing searches based on specific queries, such as keywords, authors, or topics, and then retrieves abstracts or even full texts (when available) from various academic publications. A key feature of ScholarAI is its ability to provide concise summaries of these papers. This is particularly useful for researchers who need a quick understanding of a study without reading the entire document, or even an article’s abstract. Additionally, ScholarAI includes citation management tools, which help users save and organize the most relevant papers for their research.

To illustrate the functionality of the ScholarAI plugin, we 
asked the plugin to find relevant papers on the topic of “local linear embedding”.  
\begin{figure}[h!]
    \centering
    \includegraphics[width=1.0\textwidth]{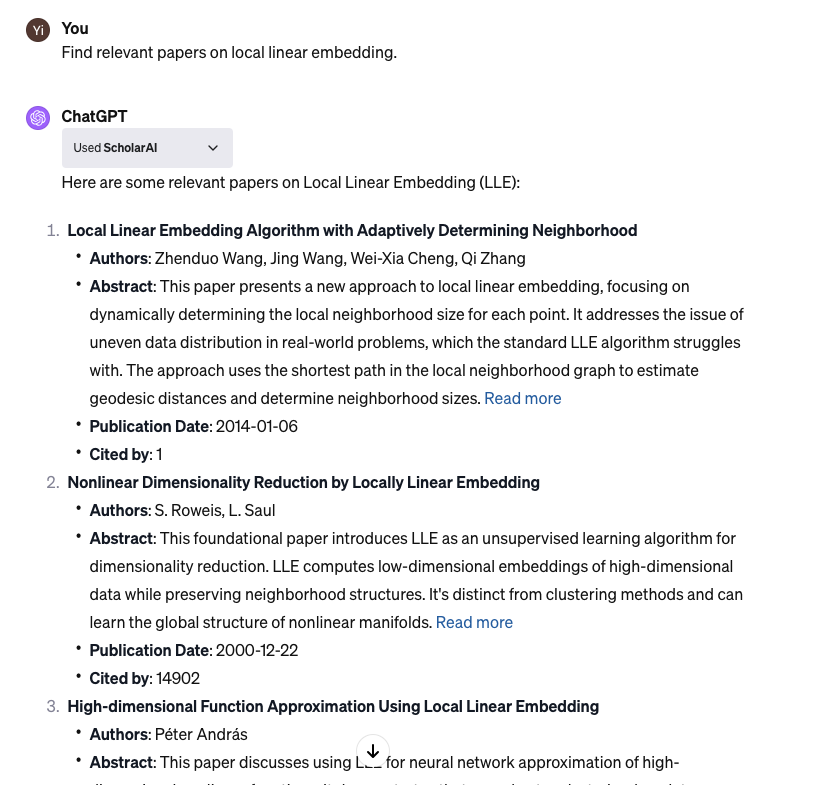}
    \caption{Search results on ``local linear embedding'' using the ScholarAI plugin.}
    \label{fig:scholarai}
\end{figure}
As shown in Figure~\ref{fig:scholarai},
the plugin returned several articles based on these keywords, each with a summary, citation counts, and source links. Notably, it included \citet{roweis2000nonlinear}, the highly cited Science paper  which was also discovered previously through the Litmaps App. ScholarAI provided a brief abstract of this paper, which allows us to quickly grasp the main ideas and importance of the paper. In addition to this key publication, the plugin also identified other relevant studies, each summarized with essential details. This enables us to get a quick understanding of each paper’s contributions without needing to read the full texts.
 
In addition to the standard ChatGPT plugins, OpenAI has recently been encouraging users to create and share their own ``custom'' GPTs in the marketplace. Since the launch of OpenAI's new GPT store, there has been a significant increase in the demand for custom GPTs across various fields and domains. Simply speaking, a custom GPT is a version of ChatGPT that is preloaded with specific knowledge and instructions for interacting with users. For example, we have noticed the recent launch of ResearchGPT (\href{https://www.researchgpt.com/}{www.researchgpt.com}), a custom GPT created by Consensus, which provides a seamless integration of ChatGPT’s conversational capabilities with access to the Semantic Scholar database of research papers, making it arguably more useful for scientific research than using the standalone Consensus website. This GPT can perform in-depth searches in academic databases, provide brief summaries and answers based on science, help write content with accurate references, and even assist in creating introductions for academic papers.
Additionally, although we utilized the ScholarAI plugin as an example above, it is important to note that OpenAI and ChatGPT are sunsetting plugins in favor of a more integrated approach. We were recently informed that ScholarAI is transitioning to a standalone GPT model, incorporating all the features of the ScholarAI plugin and more. This move is indicative of a broader trend where plugins are evolving into custom GPTs, suggesting a future where AI functionalities are seamlessly integrated directly into specialized GPT models, enhancing user experience and expanding the capabilities of generative AI in research and beyond.

\section{Summarizing and abstracting manuscripts}\label{sec:summarizing}

ChatGPT is increasingly recognized as an effective tool for summarizing research papers. In general, it does well in analyzing a whole research paper and then producing a concise summary. However, its functionality has been previously limited by the requirement for manual text input, and only recently does it support direct file uploads. ChatGPT plugin tools such as AskYourPDF and ChatwithPDF, and even ScholarAI which has been enhanced to read pdf files, were more specifically developed to enable direct handling of (multiple) pdf files, providing a more flexible option for researchers who need to engage closely with their lengthy documents and data. These tools can provide short, clear summaries of long articles, making them easier to understand. This saves time and helps to clearly show the main points of each paper, making them more digestible. 

Although ChatGPT and its associated plugins are effective for summarizing general written content, unfortunately, they encounter difficulties when it comes to more technical research papers, especially those heavy in mathematics and data analysis. The design of ChatGPT, which is more suited for natural language, can struggle to accurately summarize and abstract important details from the technical parts of these papers. This issue is particularly noticeable in areas such as statistical methodology, where crucial theorems and technical specifics are frequently presented through intricate mathematical expressions. In such cases, ChatGPT is better at offering a basic overview rather than a detailed technical analysis.

\begin{figure}[h!]
    \centering
    \includegraphics[width=1.0\textwidth]{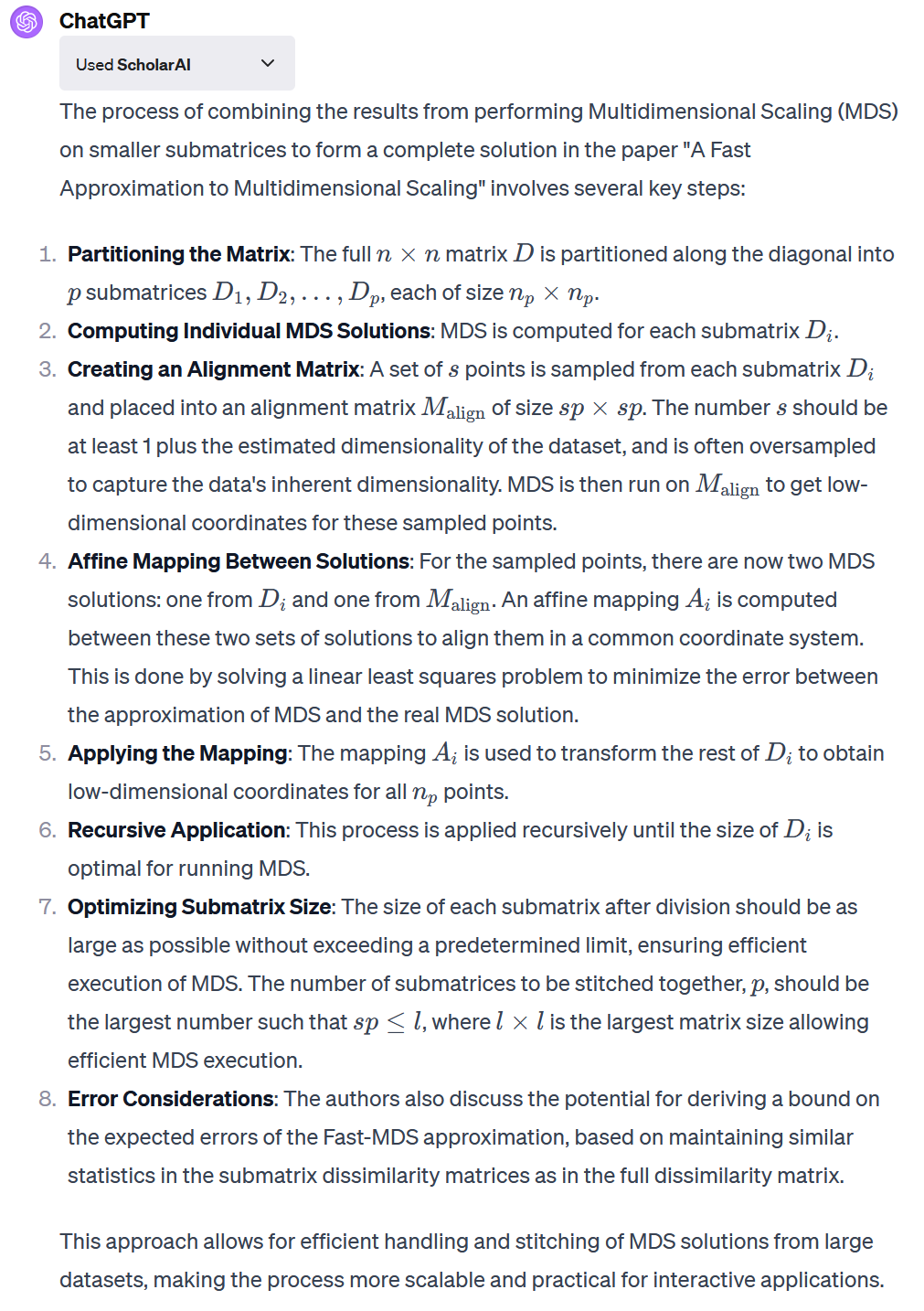}
    \caption{Summary extraction from \citet{yang2006fast}.}
    \label{fig:yang2006}
\end{figure}

As an example, we explored the accuracy of details about \citet{yang2006fast} summarized by ChatGPT.  With the ScholarAI plugin loaded, ChatGPT presented a brief summary of the initial sections of the paper.  One detail in the ChatGPT summary was “The authors propose a divide-and-conquer approach, dividing the matrix into smaller submatrices for MDS, then combining the solutions for the complete $n \times n$ matrix.”  We followed up with the question, “How are the results from performing MDS on the smaller submatrices combined to form a complete solution?”  The response by ChatGPT can be seen in Figure~\ref{fig:yang2006}. 

The answer by ChatGPT is mostly a rephrasing of the first three paragraphs of text in Section~3 
of the manuscript.  It is mostly accurate, which may not be surprising given that it is essentially a rephrasing of the body of the paper.  However, ChatGPT makes a mistake that appears in items 1 and 5 of its response.  
The method involves partitioning a dissimilarity matrix, $D$, 
into $p$ submatrices $D_1, D_2, \ldots, D_p$.
ChatGPT refers to the dimensions of the submatrices being of size $n_p\times n_p$. 
In fact, the manuscript refers to the dimensions of the 
submatrices as $\frac{n}{p}\times \frac{n}{p}$, 
so that ChatGPT misread the quotients as subscripts.  Otherwise, the technical details appear accurate.

The ability of ChatGPT to summarize research papers varies depending on the content. It currently appears to be useful for text-focused papers but less effective for those with significant technical or mathematical details.
One aspect of ChatGPT Plus (using GPT-4) that limits the size of 
research papers (or collections) it can summarize
is the context length, which is 8,000 tokens.
This is essentially the equivalent of 6,000 words of text \citep{tiwari2024}
that can be processed for analysis.
However, recent advances have allowed for much larger context lengths.
Google has recently rolled out the Gemini 1.5 Pro LLM, which
boasts a context length of about 1,000,000 tokens
(roughly 700,000 words), and has experimented successfully with
a version of Gemini that has a context length of
10,000,000 tokens \citep{pichai2024}.
This expansion in context length allows for processing documents 
of unprecedented size in a single instance, making it possible to 
analyze and synthesize information from lengthy texts such as entire books
or extensive collections of documents without needing to segment them 
into smaller pieces.
We should expect that
this technology will be accessible in the not-too-distant future.

\section{Discussion} \label{sec:discuss}

Generative AI technology has made enormous strides in such a short time that it should not be surprising that the capabilities of the tools we described are evolving rapidly.  On almost a biweekly basis, we receive email updates about new and upcoming features of the tools we reviewed in this article.  Given this dynamic landscape, attempting to encapsulate the full spectrum of features in a static publication would be futile.  Even though the state of the art in AI for use in research discovery and exploration is impressive, 
the future is even more exciting.
Insightful predictions, challenges, and possible directions for AI development have been articulated by \citet{morris2023scientists} drawing on input from scientists in various fields, and by \citet{brodnik2023perspective} in the area of mechanics.  In this context, we offer our perspective on the probable trajectories of AI advancement, aiming to further streamline and empower the processes of literature discovery and synthesis.

The tools that currently exist to discover research references are remarkable, yet the scope for enhancing the databases and repositories that underpin these research discovery tools is substantial.  
As the tools improve, larger catalogs of articles and books will
become discoverable and more readily summarized.
Tools will be developed that no longer only searched conventional formats 
such as articles and books, but will include videos (like lectures or talks), 
podcasts, slide presentations, and other non-traditional methods of 
dissemination that could greatly benefit researchers.  
This is especially valuable for those in the early stages of exploring 
fields within statistics and data science, who might prefer absorbing 
information through these alternative, more engaging formats.
With the introduction of LLMs like Google's Gemini Pro 1.5, 
which can ingest a variety of formats,
we are practically there.
However,
a significant barrier to this progress is the challenge of navigating copyright 
issues, especially when it comes to accessing literature that is not 
freely available and resides behind the paywalls of non-open access journals. 
This situation creates a bottleneck, as the most current and often highly 
innovative research is restricted, limiting the LLMs' ability to learn from 
the latest advances into their outputs.
This obstacle underscores the urgent need for innovative solutions that can 
bridge the gap between copyright restrictions and the potential for 
LLMs to contribute to research and knowledge dissemination. 
One potential pathway is the development of partnerships 
between AI developers and academic publishers, allowing for legal access 
to paywalled content for the purpose of training LLMs. 
Another solution could involve the advancement of methods for 
summarizing or paraphrasing existing research in a manner that respects 
copyright but still enables the incorporation of published research into 
the LLMs' knowledge base. 
Additionally, if open access research publication models continue to 
be adopted more widely, the problems associated with the inaccessibility
of copyrighted publications would be naturally alleviated.

In addition to expanding the resources for research discovery, the burgeoning field of extracting and synthesizing information from multiple sources is poised for considerable advancement, complementing the expansion of research discovery resources.  On a smaller scale, having reliable tools that analyze multiple documents can help to distinguish the unique contributions among them, as well as to detect common patterns or methodological trends.  On a larger scale, the ability to synthesize large collections of documents on a related topic can lead to creating review articles that would enable researchers to delve into and learn new domains within statistical scientific inquiry with greater ease and depth.

By integrating the content analysis of large collections of published research with impact measures like citation indices, it is possible to discern emerging trends of topics in statistics and data science.  
This information can guide statistical researchers in strategically choosing their future areas of focus.  
Using the same source of data, an AI tool could
determine from, say, an article’s abstract, whether the article is likely to be highly cited in the future.  
This could be achieved through careful modeling of article citation indices as a function of an abstract’s contents, providing a predictive tool for gauging the potential influence of research works.

Another type of tool that could take advantage of both large collections of published research along with AI analysis of text relates to improving citations within manuscripts.  
For example, an AI tool could analyze
a draft version of a manuscript and suggests alternative and possibly more 
foundational or highly-cited references instead of ones 
currently in the manuscript.
Even more ambitiously, 
an AI tool could analyze a draft manuscript and find instances of text that should include proper citations if they do not already exist, flagging these for the researcher's attention.  Given the amount of time researchers spend on tracking down relevant citations, off-loading this task to an AI tool would make publishing research more efficient.

Finally, the success of AI in language translation \citep{doherty2016translations, tran2018importance, gulcehre2015using} suggests its potential application in a unique academic challenge: translating specialized terminologies across various quantitative sub-fields in statistics and data science. Researchers often encounter difficulties when trying to reconcile differing terminologies used in classical statistics, computer science, machine learning, econometrics, and psychometrics. 
The development of AI tools designed to translate technical language across 
disciplines would serve as a crucial bridge, facilitating better understanding 
and integration of research across these areas. 
Such tools would also enable researchers to more easily comprehend and utilize work that might currently be challenging to interpret, thus encouraging interdisciplinary collaboration and innovation.

It is important to acknowledge that the future directions of AI in research 
discovery and summarization outlined here are unknown.
However, we believe that the foundational work for these developments is already in place, making significant advancements a matter of “when,” not “if.” The rapid pace of AI progress in recent years is nothing short of astonishing, with tangible impacts already felt by researchers in their professional lives. We anticipate that in the near future, AI tools for research tasks will have evolved to such an extent that researchers can dedicate their efforts to what truly matters: engaging in in-depth research. This shift promises to transform the research landscape, allowing researchers to focus more on innovation and less on administrative tasks.

\addtolength{\baselineskip}{-12pt}
\bibliographystyle{apacite}
\bibliography{reference}

\end{document}